\documentclass[10pt,draftclsnofoot,onecolumn,compsoc]{IEEEtran}

\ifCLASSOPTIONcompsoc
  \usepackage[nocompress]{cite}
\else
  \usepackage{cite}
\fi
\ifCLASSINFOpdf
\else
\fi

\usepackage{graphicx}
\usepackage{subfigure}
\begin{document}
%
\title{Improving Generalized Zero-Shot Learning by Semantic Discriminator}
%
%
%
%

\author{
Xinpeng~Li
\IEEEcompsocitemizethanks{\IEEEcompsocthanksitem Xinpeng Li
\protect\\
}
\thanks{*corresponding author}}

\IEEEtitleabstractindextext{%
\begin{abstract}
It is a recognized fact that the classification accuracy of unseen classes in the setting of Generalized Zero-Shot Learning (GZSL) is much lower than that of traditional Zero-Shot Leaning (ZSL). One of the reasons is that an instance is always misclassified to the wrong domain. Here  we refer to the seen and unseen classes as two domains respectively. We propose a new approach to distinguish whether the instances come from the seen  or unseen classes. First the visual feature of instance is projected into the semantic space. Then the absolute norm difference between the projected semantic vector and the class semantic embedding vector, and the minimum distance between the projected semantic vectors and the semantic embedding vectors of the seen classes are used as discrimination basis. This approach is termed as SD (Semantic Discriminator) because domain judgement of instance is performed in the semantic space. Our approach can be combined with any existing ZSL method and fully supervision classification model to form a new GZSL method.
\end{abstract}
}

\maketitle

\IEEEdisplaynontitleabstractindextext

%
\IEEEpeerreviewmaketitle

\IEEEraisesectionheading{\section{Introduction}\label{sec:introduction}}

%
%
%
%
\IEEEPARstart{T}{he} traditional fully supervised image classification task can be done very well now. However, it relies heavily on a large amount of labeled data, and the collection of labeled data often requires a lot of manpower and material resources. In order to resolve this contradiction, Zero Shot Learning (ZSL)\cite{xian2017zero,fu2018recent,lampert2009learning,jayaraman2014zero} was proposed.
In the ZSL setting, we only need to provide an auxiliary semantic embedding for each class instead of labeling each instance, which greatly reduces the workload of data preparation.
ZSL trains classification model with labeled instances from the seen classes, and then bridge the seen and unseen classes through semantic embedding.  In the traditional ZSL setting, it is worth noting that the seen  and unseen classes are completely different.

In recent years, researchers have proposed a more realistic and challenging task termed Generalized Zero-shot Learning (GZSL)\cite{chao2016empirical,xian2018feature,kumar2018generalized,zhang2017learning,liu2018generalized}. Instead of the traditional ZSL setting which only classifies the instances of unseen classes at the test stage, GZSL tries to classify the instances of both seen and unseen classes at the same time, while only the labeled instances of seen classes are provided. Since the range of classification is the set of all the seen and unseen classes,  the task of GZSL is more difficult.
The existing GZSL methods can be roughly divided into two types. The first kind of methods are similar to most of traditional ZSL methods \cite{xian2018feature,felix2018multi,kumar2018generalized,zhu2018generative,sung2018learning,zhang2017learning,guo2018implicit,zhang2018triple,liu2018generalized,socher2013zero}, which aim to construct a model so that it can perform well on both the seen and unseen classes. This approach usually tends to sacrifice the classification accuracy of instances of seen classes.

Another kind of methods\cite{socher2013zero,chao2016empirical,liu2018generalized,atzmon2019adaptive} usually consist of three parts. In the first part, given an instance, a binary classification module is used to determine which domain it is from. Then, based on the prediction result, a fully supervised classification model trained by seen classes is used  if it is predicted to belong to the seen classes. Otherwise, a traditional ZSL model is used.  Our method also belongs to this category. The benefit of this research route is that the traditional fully supervised classification and ZSL models are relatively mature and doing these tasks separately is much easier.

The key issue in the second kind of methods lies in how to accurately judge an instance which domain it is from  as far as possible, whose accuracy will directly affect the overall classification performance. Most existing methods use the distribution of prediction scores of the instances in various categories to determine the domain. However, because the instances from the unseen classes have not been used in the training of classification model, the classification scores of  instances of unseen classes are rather scattered. There always exist some instances whose scores are very close to those of instances of seen classes. Thus these instances  will  be misclassified.

Instead of the above approaches, we find some interesting phenomena that are very helpful to distinguish whether the instance is from the seen or unseen classes. We project the instances of seen classes into the semantic space and align these projected semantic vectors with the corresponding class semantic embedding vectors. It can be observed that the norm length of the projected semantic vectors is always around the norm length  of the semantic embedding vectors. It is not strange because the semantic embedding vectors are always been normalized to be the same. When the instances of unseen classes are projected into the semantic space through the learned mapping trained by the seen classes, it can be found that the norm length of the projected semantic vectors will be relatively far away from the norm length of the semantic embedding vectors. At the same time, we also find that the minimum distance between the projected semantic vectors of instances of seen classes and the semantic embedding vectors of  seen classes are always kept at a very low value;  however, for the instances of unseen classes, the minimum distances are always larger.

Based on the above mentioned phenomena, we put forward several simple and efficient strategies to determine which domain an instance belongs to.  For the first strategy, we project the instance into the semantic space and compute the norm  difference with the semantic embedding vectors. If the absolute difference is below to a threshold, the instance will be from the seen classes; otherwise it will be from the unseen classes. Furthermore, when the minimum distance between the projected semantic vector and the semantic embedding vectors of the seen classes is small, the instance will be considered to be from the seen classes; otherwise, it will belongs to the unseen classes. By combining the first strategy, the second strategy is formed to improve the distinguish performance. Naturally, the first and second strategies can be combined to form a new strategy which can obtain middle performance. The overall approach is termed Semantic Discriminator (SD), because we use the relationship between the projected semantic vectors and the semantic embedding vectors to judge which domain  an instance is from. On this basis, we can combine our approach with any existing ZSL method and fully supervised classification model to complete the task of GZSL. Our contributions can be summarized as follows.

1) Some interesting phenomena are observed. After the instance  is projected into the semantic space, the norm difference with the semantic embedding vectors are different respect to which domain the instance belongs to. Furthermore, the minimum distance between the projected semantic vector of  instance of seen classes and the semantic embedding vectors of seen classes are always small.

2) Based on these observations, we propose a new approach (SD) to judge whether  an instance is from the seen or unseen classes. It consists of several strategies and is very simple and easy to be implemented without artificial parameter setting.
By combining any existing ZSL and fully supervised classification model, the GZSL task can be done.

\begin{figure*}[t]
  \centering
  \includegraphics[scale=0.43]{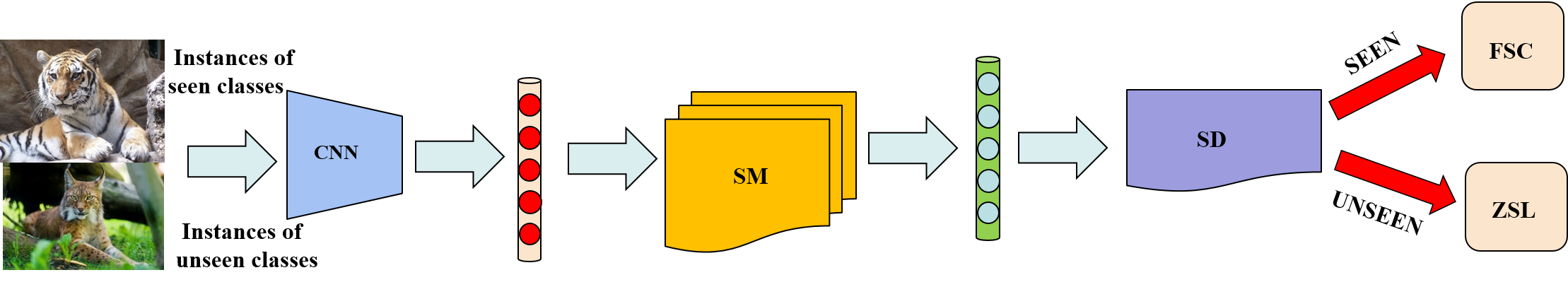}
  \caption{The overall framework of our approach (SDGZSL). SD(Semantic Discriminator) is a discriminator, which is used to determine which domain an instance comes from. If the instance belongs to the seen classes, we uses FSC (a Fully Supervised Classification model) to further classify it. Otherwise, the instance are sent to a ZSL model for further classification.}
  \label{fig:pathdemo4}
\end{figure*}

\section{Related work}
Both ZSL and GZSL use semantic embedding as a bridge to transfer the knowledge learned from the seen classes to the unseen classes.  The semantics we usually use are attribute\cite{lampert2009learning} and word vectors, etc. The attribute is a detailed introduction to certain kind of things. For example, the attributes in the animal dataset AWA are such as: black, hairless, longleg, etc. By learning how to extract these attributes from the images in the seen classes, we can use this knowledge in the unseen classes to extract the corresponding attributes to match the known unseen attribute to complete the classification task.
The word vectors can be got by a language model, such as Word2vec\cite{reed2016learning,zhu2018generative,socher2013zero} and Glove which learn the relationship between each word from a large-scale corpus so that a word corresponds to a vector.
In ZSL or GZSL, we can use the word vectors of the class name as the semantic embedding.

The existing GZSL mainly uses two kinds of methods. The first and now mainstream kind of methods try to build a model so that it can have good performance in both the seen and unseen classes. This kind of methods can be divided into two major categories, one of which is the generative methods that use GAN\cite{goodfellow2014generative}, VAE, etc, which generates visual features through the unseen semantic embedding vectors, and uses these features for training classification models for unseen classes\cite{xian2018feature,felix2018multi,kumar2018generalized,zhu2018generative}. Another category of methods are to bridge the visual and semantic information directly by learning the mapping between the visual  and semantic information\cite{sung2018learning,zhang2017learning,guo2018implicit,zhang2018triple,liu2018generalized,socher2013zero}.

The remaining kind of GZSL methods try to determine which domain an given instance is from. The advantage of this kind of methods is that knowing which domain an instance belongs to will reduce disturbances from another domain.  So how to accurately distinguish between the instances of seen and unseen classes has become the most critical and challenging problem in this kind of methods. There exist some methods. For example, \cite{socher2013zero} uses a hard gating mechanism to predict which domain an instance belongs.The hard gating with some fixed parameters can not make the corresponding adjustments to the models obtained in different situations. \cite{chao2016empirical} calibrates the seen and unseen scores by subtracting one of the seen scores to distinguish the instances by these scores. The diversity of instances of unseen classes is not fully considered. \cite{liu2018generalized} uses temperature scaling and entropy regularization functions to distinguish.
COSMO\cite{atzmon2019adaptive} uses a soft gating by building an out-of-distribution detector. Although these methods have somewhat alleviated the problems we mentioned above,  they are more complicated to be implemented and the performance is not particularly very well. Our approach is simple and dynamically obtains parameters from statistical results for better generalization.

Our approach is inspired by the work in \cite{xu2019larger}. It has shown that the norm lengths of projected instances from the source and the target domains are significantly different, i.e. the norm lengths of instances in the target domain will be smaller. While the GZSL task is rather different from this case. The norm length of the projected semantic vectors of  instances of unseen classes are not small. In this paper, we propose two new observations  which are very simple to be implemented and can greatly improve the performance of GZSL task.

\section{The Proposed Approach}
\subsection{Problem Definition}
Assume a source dataset \(D_s=\{(x_i,y_i,z_i)\}_{i=1}^{N_s}\), where \(x_i{\in}X_s{\subset}R^{d{\times}N_s}\) is the $i$th instance in the seen classes,  \(N_s\) is the cardinality  of $D_s$, \(y_i{\in}Y_s{\subset}R^{C_s{\times}N_s}\) is the corresponding labels and \(z_i{\in}Z_s{\subset}R^{S{\times}C_s}\) is the corresponding semantic embedding, where \(C_S\) is the number of catagories in the source dataset. A target dataset \(D_u\) consists of two parts: \(X_u=\{x_j\}_{j=1}^{N_u}{\subset}R^{d{\times}N_u}\) is the set of unlabeled instances, and \(Y_u{\subset}R^{C_u{\times}N_u}\) is the corresponding set of labels and which are unknown. \(Z_u{\subset}R^{S{\times}C_u}\) is the set of attributes of unseen classes.

Since the norm lengths of all semantic embedding vectors are unified to the same in advance, we record it as \(l\). For the GZSL task, the goal is to learn a function \(C:\{X_s,X_u\}{\to}Y_s{\cup}Y_u\), where \(Y_s{\cap}Y_u={\emptyset}\).

\subsection{Overall Idea}
The overall framework of our approach is shown in Fig. 2. The proposed method includes four modules: Semantic Mapping (SM) module, Semantic Discriminator (SD) module, a Fully Supervised Classification (FSC) module and a ZSL module.

Given an instance, we first use a CNN to extract the visual feature and then use SM to project this visual feature into the semantic space to get a projected semantic vector, and then feed the projected semantic vector into SD. If SD determines that this instance belongs to the seen classes, the feature of this instance will be sent to FSC for further classification. Otherwise, this instance will be sent to the ZSL module for further classification. We will illustrate these four modules in detail below.

\subsection{Semantic Mapping Module}
Given an instance,  a pre-trained convolutional neural network is first used to perform preliminary feature extraction. We use resnet101 \cite{he2016deep} to do this job. The preliminary extracted features are then feed into a multilayer perceptron (MLP). This MLP is called Semantic Mapping(SM) module, and supervised trained by instances of seen classes and their corresponding semantic embedding vectors. The corresponding loss function is defined as follows,
\begin{equation}
L_{SM}=\frac{1}{N_s}\sum_{i=1}^{N_s} \left \|F(g_c(x_i^s))-z_i^s \right\|_2^2
\end{equation}
where the \(F\) represents the mapping corresponding to SM and \(g_c\) is the mapping corresponding to the CNN. \(x_i^s\) represents the $i$-th image instance  of seen classes, \(z_i^s\) is the corresponding semantic embedding vector and \(N_s\) is the number of instances of seen classes.
\subsection{Semantic Discriminator}
Based on the observations introduced in Section 1, we propose three strategies each of which has its own advantages. Next, we will introduce these three strategies in detail.

The first strategy only use the norm length of the projected semantic vector as a basis for judgment. At first we define \(D_{l}\) as the absolute norm difference between the projected semantic vector and the unified norm \(l\) respect to all semantic embedding vectors,
\begin{equation}
D_{l}(x)=| \left \|F(g_c(x))\right \|_2-l |.
\end{equation}
As denoted before we found that most of \(D_{l}\) of seen classes are very small;  while it is usually larger for the instances of unseen classes.

Therefore, we set a threshold. If the \(D_{l}\) of an instance is below this threshold, it is considered to be from the seen classes, otherwise it is  from the unseen classes. We call this strategy as Semantic Discriminator Only by Length (\(SD_{OL}\)), which can be expressed as the following,
\begin{equation}
SD_{OL}(x)=\left\{
\begin{array}{lll}
S,       &      & {D_{l}(x)<R_{OL}},\\
U,     &      & {D_{l}(x){\ge}R_{OL}}
\end{array} \right.
\end{equation}
where S and U indicate that the instance is from the seen classes or unseen classes respectively. The threshold \(R_{OL}\) is a parameter, which can be obtained by the mean and standard deviation of the \(D_l\) of seen classes,
\begin{eqnarray}
R_{OL}=m_{D_l}+var_{D_l}
\end{eqnarray}
where
\begin{eqnarray*}
m_{D_l}&=&\frac{1}{N_s}\sum_{i=1}^{N_s} D_{l}(x_i^s),\\
var_{D_l}&=&\sqrt{\frac{1}{N_s}\sum_{i=1}^{N_s} (D_{l}(x_i^s)-m_{D_l})^2}.
\end{eqnarray*}
Here, \(x_i^s\) represents the $i$-th instance of seen classes. The reason we get the threshold value in this way is that  the \(D_l\) of the seen classes obeys the normal distribution since we force to align the projected semantic vectors with the semantic embedding vectors. For the normal distribution, most of the elements are within the standard deviation from the mean. So we let the threshold value decided by seen classes adaptively.
In this way, the parameter  has better applicability to different models. For the same reason, we can obtain the thresholds mentioned later in a similar way.

Although \(SD_{OL}\) has already achieved good results, there are still some instances that will be misclassified.
Fortunately, we found that the distances between the projected semantic vector and the seen semantic embedding vectors are regular.
If an instance is from the seen classes, then there will exist  a very small distance; otherwise these distances are all usually large. We refer to the minimum  distance between the projected semantic vector and the semantic embedding vectors of seen classes as the Minimum Semantic Distance (MSD). It can be calculated as the following,
\begin{equation}
MSD(x)=\mathop{min}_{z{\in}Z_s} \ \|F(x)-z\|_2^2,
\end{equation}
where $x$ is any instance and $z\in Z_s$ is the corresponding semantic embedding vector of seen classes.

Unfortunately, this rule cannot be used alone to determine which domain an instance comes from since the distance distribution between the projected semantic vector and the semantic embedding vectors of unseen classes is not clear.   So we combine this rule with \(SD_{OL}\) to propose a second strategy. For the instance misclassified as seen classes by  \(SD_{OL}\), if MSD is large, it can be excluded from the seen classes. On the other hand,  for the instance misclassified as unseen classes, if MSD is small, it can be included in the seen classes. This strategy is referred to as  Semantic Discriminator through Minimum Distance After the Length discrepancy (\(SD_{DL}\)), which is expressed as
\begin{equation}
SD_{DL}(x)=\left\{
\begin{array}{lll}
S, & {D_{l}(x)<R_{OL}\cap MSD(x)<R_0},\\
S, & {D_{l}(x){\ge}R_{OL}\cap MSD(x)<R_1},\\
U, & {D_{l}(x)<R_{OL}\cap MSD(x){\ge}R_0},\\
U, & {D_{l}(x){\ge}R_{OL}\cap MSD(x){\ge}R_1}.
\end{array} \right.
\end{equation}
For all instances, we first use \(SD_{OL}\) for classification, and then set two thresholds \(R_0\) and \(R_1\). For the instance that is classified to the seen classes by \(SD_{OL}\), if its MSD is less than \(R_0\), the prediction is reinforced that it belongs to the seen classes; otherwise it is considered to belong to unseen classes. On the other hand, if both \(SD_{OL}\) and MSD are larger than the thresholds, the prediction to be unseen classes will be reinforced. Otherwise, if MSD less than \(R_1\) and \(SD_{OL}\) is larger than \(R_{OL}\), then the prediction will be corrected to be the seen classes.

Here, two parameters are defined as follows,
\begin{eqnarray}
R_0&=&m_{MSD}+2std_{MSD}, \\
R_1&=&m_{MSD}+std_{MSD}
\end{eqnarray}
where
\begin{eqnarray*}
m_{MSD}&=&\frac{1}{N_s}\sum_{i=1}^{N_s} MSD(x_i^s),\\
std_{MSD}&=&\sqrt{\frac{1}{N_s}\sum_{i=1}^{N_s} (MSD(x_i^s)-m_{MSD})^2}
\end{eqnarray*}
and $x_i^s$ is the $i$th instance of the seen classes. Here we set two different thresholds because the prediction we make here is based on the results of \(SD_{OL}\). For the instances that are decided by \(SD_{OL}\) to belong to the seen classes, there is a high confidence. Therefore, we should relax the standard  to re-determine the results by using MSD.  For the instances that are identified as unseen classes, the probability that they belong to the seen classes is very small.  When their MSD are small enough, they will belong to the seen classes. So a smaller threshold should be set for them.

In the two strategies proposed above, \(SD_{OL}\) is simpler and faster, but \(SD_{DL}\) achieves better results. Based on the advantages and disadvantages of these two methods, we propose the third strategy, which we call Semantic Discriminator through Weighted Sum of Length difference and Minimum Distance (\(SD_{WS}\)). Given an instance, we first get the weighted sum of its length diffence and its MSD and then set a threshold \(R_{WS}\). If its weighted sum is less than \(R_{WS}\), it is considered to be from the seen classes. Otherwise it is considered to belong to the unseen classes. It can be expressed as
\begin{equation}
SD_{WS}(x)=\left\{
\begin{array}{lll}
S,&&{D_{l}(x)+{\lambda}MSD(x)<R_{WS}},\\
U,&&{D_{l}(x)+{\lambda}MSD(x){\ge}R_{WS}}
\end{array} \right.
\end{equation}
where \(\lambda\) is a tunable parameter, we set it to 1. And the \(R_{WS}\) is a parameter which can be calculated as

\begin{equation}
R_{WS}=m_{WS}+std_{WS}
\end{equation}
where
\begin{eqnarray*}
m_{WS}&=&\frac{1}{N_s}\sum_{i=1}^{N_s}   (D_{l}(x_i^s)+{\lambda}MSD(x_i^s)),\\
std_{WS}&=&\sqrt{\frac{1}{N_s}\sum_{i=1}^{N_s} (D_{l}(x_i^s)+{\lambda}MSD(x_i^s)-m_{WS})^2}.
\end{eqnarray*}
Theoretically, the third strategy  can achieve intermediate performance between the first and second strategies.

\subsection{Fully Supervised Classification Module}
The instance belonging to the seen classes judged by SD will be further classified  through FSC, and its predicted semantic embedding is
\begin{equation}
f_s(x)=\mathop{\arg\max}_{z{\in}Z_s} \ f_F(g_c(x),z).
\end{equation}
where \(f_F\) is any fully supervised classification model, and the output  is the probability that the instance belongs to a certain seen class. We know that each class corresponds to a unique semantic embedding. So when a semantic embedding  corresponds to this instance, we also know the class label of this instance.

\subsection{ZSL Module}
The instance belonging to the unseen classes will be given a further prediction through an existing ZSL method, and its predicted semantic embedding is
\begin{equation}
f_u(x)=\mathop{\arg\max}_{z{\in}Z_u} \ ZSL(g_c(x),z),
\end{equation}
where the output of ZSL is the probability that the instance belongs to a certain unseen class.
\section{Experiments}
\subsection{Datasets}
In order to verify the effectiveness of our method, we perform a large number of experiments on 4 datasets widely used in the GZSL research literatures. They are AWA, CUB, aPY and SUN.

\textbf{AWA}\cite{xian2017zero}. Animals with Attributes (AWA) includes 50 animal classes with a total of 30,475 images. Of these, 40 classes are the seen and 10 classes are the unseen. This dataset provides 85-dimensional attribute vectors as semantic embeddings.

\textbf{CUB}\cite{wah2011caltech}. It is a dataset for fine-grained classification. It provides a total of 11,788 pictures from 200 classes. Of these, 150 classes are seen, and the remaining 50 classes are unseen. It uses 312-dimensional attribute vectors for semantic embeddings.

\textbf{aPY}. aPascal-aYahoo(aPY) includes 15339 images from 32 classes. We use 20 classes as seen classes, and the remaining 12 classes as the unseen classes. Its semantic embedding is described as 64-dimensional attributes.

\textbf{SUN}\cite{patterson2012sun}. It is a dataset for classification of complex visual scenes. It contains 14,340 images from 717 categories. Of these, 635 categories are used as seen classes, and the other 72 categories are used as unseen classes. At the same time, it provides 102-dimensional attribute vector as semantic embedding.

\subsection{Evaluation Metrics}
For GZSL we need to comprehensively consider the performance of the seen and unseen classes, so we calculate the harmonic mean of accuracies as the previous work\cite{xian2017zero},
\begin{equation}
  H_{acc}=\frac{2{\times}acc_S{\times}acc_{U}}{acc_{S}+acc_{U}}
\end{equation}
where \(acc_ {S}\) and \(acc_ {U}\) are the top-1 accuracies of the instances of seen and unseen classes respectively. The top-1 accuracy is calculated as follows,
\begin{equation}
  acc_Y=\frac{1}{|Y|}\sum_{c=1}^{|Y|}\frac{\#correct\ predictions\ in\ c}{\#samples\ in\ c}
  \end{equation}

where \(|Y|\) is the total number of corresponding classes.

\section{Conclusion}
We proposed a classifier named Semantic Discriminator for distinguishing whether an instance comes from the seen  or  unseen classes. Unlike other GZSL methods, which use the scores of instances on various classes as a basis or train a network to adaptively give judgments, our approach directly uses some inherent information of the projected semantic features to identify. At the same time, our method is simple and free of complex parameter setting.

\ifCLASSOPTIONcaptionsoff
  \newpage
\fi


\begin{thebibliography}{1}
\bibitem{akata2015evaluation}
Z.~Akata, S.~Reed, D.~Walter, H.~Lee, and B.~Schiele, ``Evaluation of output
  embeddings for fine-grained image classification,'' in \emph{Proceedings of
  the IEEE Conference on Computer Vision and Pattern Recognition}, 2015, pp.
  2927--2936.

\bibitem{akata2015label}
Z.~Akata, F.~Perronnin, Z.~Harchaoui, and C.~Schmid, ``Label-embedding for
  image classification,'' \emph{IEEE transactions on pattern analysis and
  machine intelligence}, vol.~38, no.~7, pp. 1425--1438, 2015.

\bibitem{atzmon2019adaptive}
Y.~Atzmon and G.~Chechik, ``Adaptive confidence smoothing for generalized
  zero-shot learning,'' in \emph{Proceedings of the IEEE Conference on Computer
  Vision and Pattern Recognition}, 2019, pp. 11\,671--11\,680.

\bibitem{changpinyo2016synthesized}
S.~Changpinyo, W.-L. Chao, B.~Gong, and F.~Sha, ``Synthesized classifiers for
  zero-shot learning,'' in \emph{Proceedings of the IEEE conference on computer
  vision and pattern recognition}, 2016, pp. 5327--5336.

\bibitem{chao2016empirical}
W.-L. Chao, S.~Changpinyo, B.~Gong, and F.~Sha, ``An empirical study and
  analysis of generalized zero-shot learning for object recognition in the
  wild,'' in \emph{European Conference on Computer Vision}.\hskip 1em plus
  0.5em minus 0.4em\relax Springer, 2016, pp. 52--68.

\bibitem{felix2018multi}
R.~Felix, V.~B. Kumar, I.~Reid, and G.~Carneiro, ``Multi-modal cycle-consistent
  generalized zero-shot learning,'' in \emph{Proceedings of the European
  Conference on Computer Vision (ECCV)}, 2018, pp. 21--37.

\bibitem{frome2013devise}
A.~Frome, G.~S. Corrado, J.~Shlens, S.~Bengio, J.~Dean, M.~Ranzato, and
  T.~Mikolov, ``Devise: A deep visual-semantic embedding model,'' in
  \emph{Advances in neural information processing systems}, 2013, pp.
  2121--2129.

\bibitem{fu2018recent}
Y.~Fu, T.~Xiang, Y.-G. Jiang, X.~Xue, L.~Sigal, and S.~Gong, ``Recent advances
  in zero-shot recognition: Toward data-efficient understanding of visual
  content,'' \emph{IEEE Signal Processing Magazine}, vol.~35, no.~1, pp.
  112--125, 2018.

\bibitem{goodfellow2014generative}
I.~Goodfellow, J.~Pouget-Abadie, M.~Mirza, B.~Xu, D.~Warde-Farley, S.~Ozair,
  A.~Courville, and Y.~Bengio, ``Generative adversarial nets,'' in
  \emph{Advances in neural information processing systems}, 2014, pp.
  2672--2680.

\bibitem{guo2018implicit}
Y.~Guo, G.~Ding, J.~Han, S.~Zhao, and B.~Wang, ``Implicit non-linear similarity
  scoring for recognizing unseen classes.'' in \emph{IJCAI}, 2018, pp.
  4898--4904.

\bibitem{he2016deep}
K.~He, X.~Zhang, S.~Ren, and J.~Sun, ``Deep residual learning for image
  recognition,'' in \emph{Proceedings of the IEEE conference on computer vision
  and pattern recognition}, 2016, pp. 770--778.

\bibitem{jacobs1991adaptive}
R.~A. Jacobs, M.~I. Jordan, S.~J. Nowlan, and G.~E. Hinton, ``Adaptive mixtures
  of local experts,'' \emph{Neural computation}, vol.~3, no.~1, pp. 79--87,
  1991.

\bibitem{jayaraman2014zero}
D.~Jayaraman and K.~Grauman, ``Zero-shot recognition with unreliable
  attributes,'' in \emph{Advances in neural information processing systems},
  2014, pp. 3464--3472.

\bibitem{kodirov2017semantic}
E.~Kodirov, T.~Xiang, and S.~Gong, ``Semantic autoencoder for zero-shot
  learning,'' in \emph{Proceedings of the IEEE Conference on Computer Vision
  and Pattern Recognition}, 2017, pp. 3174--3183.

\bibitem{lampert2009learning}
C.~H. Lampert, H.~Nickisch, and S.~Harmeling, ``Learning to detect unseen
  object classes by between-class attribute transfer,'' in \emph{2009 IEEE
  Conference on Computer Vision and Pattern Recognition}.\hskip 1em plus 0.5em
  minus 0.4em\relax IEEE, 2009, pp. 951--958.

\bibitem{lampert2013attribute}
C.~H. Lampert, H.~Nickisch, and S.~Harmeling, ``Attribute-based classification
  for zero-shot visual object categorization,'' \emph{IEEE transactions on
  pattern analysis and machine intelligence}, vol.~36, no.~3, pp. 453--465,
  2013.


\bibitem{li2019leveraging}
J.~Li, M.~Jing, K.~Lu, Z.~Ding, L.~Zhu, and Z.~Huang, ``Leveraging the
  invariant side of generative zero-shot learning,'' in \emph{Proceedings of
  the IEEE Conference on Computer Vision and Pattern Recognition}, 2019, pp.
  7402--7411.

\bibitem{liu2018generalized}
S.~Liu, M.~Long, J.~Wang, and M.~I. Jordan, ``Generalized zero-shot learning
  with deep calibration network,'' in \emph{Advances in Neural Information
  Processing Systems}, 2018, pp. 2005--2015.

\bibitem{norouzi2013zero}
M.~Norouzi, T.~Mikolov, S.~Bengio, Y.~Singer, J.~Shlens, A.~Frome, G.~S.
  Corrado, and J.~Dean, ``Zero-shot learning by convex combination of semantic
  embeddings,'' \emph{arXiv preprint arXiv:1312.5650}, 2013.

\bibitem{patterson2012sun}
G.~Patterson and J.~Hays, ``Sun attribute database: Discovering, annotating,
  and recognizing scene attributes,'' in \emph{2012 IEEE Conference on Computer
  Vision and Pattern Recognition}.\hskip 1em plus 0.5em minus 0.4em\relax IEEE,
  2012, pp. 2751--2758.

\bibitem{reed2016learning}
S.~Reed, Z.~Akata, H.~Lee, and B.~Schiele, ``Learning deep representations of
  fine-grained visual descriptions,'' in \emph{Proceedings of the IEEE
  Conference on Computer Vision and Pattern Recognition}, 2016, pp. 49--58.

\bibitem{romera2015embarrassingly}
B.~Romera-Paredes and P.~Torr, ``An embarrassingly simple approach to zero-shot
  learning,'' in \emph{International Conference on Machine Learning}, 2015, pp.
  2152--2161.

\bibitem{schonfeld2019generalized}
E.~Schonfeld, S.~Ebrahimi, S.~Sinha, T.~Darrell, and Z.~Akata, ``Generalized
  zero-and few-shot learning via aligned variational autoencoders,'' in
  \emph{Proceedings of the IEEE Conference on Computer Vision and Pattern
  Recognition}, 2019, pp. 8247--8255.

\bibitem{shazeer2017outrageously}
N.~Shazeer, A.~Mirhoseini, K.~Maziarz, A.~Davis, Q.~Le, G.~Hinton, and J.~Dean,
  ``Outrageously large neural networks: The sparsely-gated mixture-of-experts
  layer,'' \emph{arXiv preprint arXiv:1701.06538}, 2017.

\bibitem{socher2013zero}
R.~Socher, M.~Ganjoo, C.~D. Manning, and A.~Ng, ``Zero-shot learning through
  cross-modal transfer,'' in \emph{Advances in neural information processing
  systems}, 2013, pp. 935--943.

\bibitem{sung2018learning}
F.~Sung, Y.~Yang, L.~Zhang, T.~Xiang, P.~H. Torr, and T.~M. Hospedales,
  ``Learning to compare: Relation network for few-shot learning,'' in
  \emph{Proceedings of the IEEE Conference on Computer Vision and Pattern
  Recognition}, 2018, pp. 1199--1208.

\bibitem{verma2017simple}
V.~K. Verma and P.~Rai, ``A simple exponential family framework for zero-shot
  learning,'' in \emph{Joint European Conference on Machine Learning and
  Knowledge Discovery in Databases}.\hskip 1em plus 0.5em minus 0.4em\relax
  Springer, 2017, pp. 792--808.

\bibitem{kumar2018generalized}
V.~Kumar~Verma, G.~Arora, A.~Mishra, and P.~Rai, ``Generalized zero-shot
  learning via synthesized examples,'' in \emph{Proceedings of the IEEE
  conference on computer vision and pattern recognition}, 2018, pp. 4281--4289.

\bibitem{wah2011caltech}
C.~Wah, S.~Branson, P.~Welinder, P.~Perona, and S.~Belongie, ``The caltech-ucsd
  birds-200-2011 dataset,'' 2011.

\bibitem{xian2016latent}
Y.~Xian, Z.~Akata, G.~Sharma, Q.~Nguyen, M.~Hein, and B.~Schiele, ``Latent
  embeddings for zero-shot classification,'' in \emph{Proceedings of the IEEE
  Conference on Computer Vision and Pattern Recognition}, 2016, pp. 69--77.

\bibitem{xian2017zero}
Y.~Xian, B.~Schiele, and Z.~Akata, ``Zero-shot learning-the good, the bad and
  the ugly,'' in \emph{Proceedings of the IEEE Conference on Computer Vision
  and Pattern Recognition}, 2017, pp. 4582--4591.

\bibitem{xian2018feature}
Y.~Xian, T.~Lorenz, B.~Schiele, and Z.~Akata, ``Feature generating networks for
  zero-shot learning,'' in \emph{Proceedings of the IEEE conference on computer
  vision and pattern recognition}, 2018, pp. 5542--5551.

\bibitem{xu2019larger}
R.~Xu, G.~Li, J.~Yang, and L.~Lin, ``Larger norm more transferable: An adaptive
  feature norm approach for unsupervised domain adaptation,'' in
  \emph{Proceedings of the IEEE International Conference on Computer Vision},
  2019, pp. 1426--1435.

\bibitem{yuksel2012twenty}
S.~E. Yuksel, J.~N. Wilson, and P.~D. Gader, ``Twenty years of mixture of
  experts,'' \emph{IEEE transactions on neural networks and learning systems},
  vol.~23, no.~8, pp. 1177--1193, 2012.

\bibitem{zhang2015zero}
Z.~Zhang and V.~Saligrama, ``Zero-shot learning via semantic similarity
  embedding,'' in \emph{Proceedings of the IEEE international conference on
  computer vision}, 2015, pp. 4166--4174.

\bibitem{zhang2017learning}
L.~Zhang, T.~Xiang, and S.~Gong, ``Learning a deep embedding model for
  zero-shot learning,'' in \emph{Proceedings of the IEEE Conference on Computer
  Vision and Pattern Recognition}, 2017, pp. 2021--2030.

\bibitem{zhang2018triple}
H.~Zhang, Y.~Long, Y.~Guan, and L.~Shao, ``Triple verification network for
  generalized zero-shot learning,'' \emph{IEEE Transactions on Image
  Processing}, vol.~28, no.~1, pp. 506--517, 2018.

\bibitem{zhu2018generative}
Y.~Zhu, M.~Elhoseiny, B.~Liu, X.~Peng, and A.~Elgammal, ``A generative
  adversarial approach for zero-shot learning from noisy texts,'' in
  \emph{Proceedings of the IEEE conference on computer vision and pattern
  recognition}, 2018, pp. 1004--1013.

\bibitem{zhu2019learning}
Y.~Zhu, J.~Xie, B.~Liu, and A.~Elgammal, ``Learning feature-to-feature
  translator by alternating back-propagation for generative zero-shot
  learning,'' in \emph{Proceedings of the IEEE International Conference on
  Computer Vision}, 2019, pp. 9844--9854.


\end{thebibliography}
\end{document}